\documentclass[letterpaper, 10 pt, conference]{ieeeconf}  

\IEEEoverridecommandlockouts                              

\renewcommand{\baselinestretch}{0.96} 
\usepackage{booktabs}
\usepackage{times}

\usepackage{multicol}
\usepackage[bookmarks=true]{hyperref}
\usepackage{amsmath}
\usepackage{amsfonts}
\usepackage{{booktabs}}
\usepackage{bm}
 \usepackage{array,multirow,graphicx}

\usepackage{color}

\newcommand{\vtxt}[1]{{{\color{red} V: #1}}}
\newcommand{\mtxt}[1]{{{\color{green} M: #1}}}

\newcommand{\commentout}[1]{}
\newcommand{\ourmethod}{GG-LLM}

\usepackage{balance} 

\usepackage{setspace} 
\usepackage{dblfloatfix}
 
\title{\LARGE \bf
GG-LLM: Geometrically Grounding Large Language Models for Zero-shot Human Activity Forecasting in Human-Aware Task Planning}

\author{Moritz A. Graule$^{1}$ and Volkan Isler$^{1}$
\thanks{$^{1}$Samsung AI Center New York, New York, NY 10010, USA
        {\tt\small \{m.graule,ibrahim.i\}@samsung.com}}%
}

\begin{document}

\maketitle
\thispagestyle{empty}
\pagestyle{empty}

\begin{abstract}

A robot in a human-centric environment needs to account for the human's intent and future motion in its task and motion planning to ensure safe and effective operation. This requires symbolic reasoning about probable future actions and the ability to tie these actions to specific locations in the physical environment. While one can train behavioral models capable of predicting human motion from past activities, this approach requires large amounts of data to achieve acceptable long-horizon predictions. More importantly, the resulting models are constrained to specific data formats and modalities. Moreover, connecting predictions from such models to the environment at hand to ensure the applicability of these predictions is an unsolved problem. 
We present a system that utilizes a Large Language Model (LLM) to infer a human's next actions from a range of modalities without fine-tuning. A novel aspect of our system that is critical to robotics applications is that it links the predicted actions to specific locations in a semantic map of the environment. Our method leverages the fact that LLMs, trained on a vast corpus of text describing typical human behaviors, encode substantial world knowledge, including probable sequences of human actions and activities. 
We demonstrate how these localized activity predictions can be incorporated in a human-aware task planner for an assistive robot to reduce the occurrences of undesirable human-robot interactions by $29.2\%$ on average. 

\end{abstract}

\IEEEpeerreviewmaketitle

\section{Introduction}


As robots increasingly move to settings such as warehouses, hospitals, and our homes, they will be required to dynamically operate in proximity to, and collaboration with, human operators. Mere reactive collision-avoidance is not always sufficient to achieve safe and effective assistance in these scenarios. Rather, robots must possess the ability to proactively infer the human operators' intentions and predict their future actions and trajectories. They must further be able to leverage their predictions to create human-aware task and motion plans that maximize the robots' performance while minimizing undesirable interactions with humans (like collisions, uncomfortable proximity, or disturbing audio). The creation of these plans requires the integration of symbolic reasoning to predict human actions and the localization of these predictions within the physical environment. 

Prior work has demonstrated impressive capabilities in human activity recognition and anticipation through training specific models on data from smartphones and wearable sensors~\cite{ramanujam2021human}, video streams~\cite{rajasegaran2023benefits}, as well as sensor-equipped homes~\cite{van2008accurate,tapia2004activity}. However, there remain a number of challenges with these approaches, including the substantial data required for training, the inherent confinement to the training modality, the limited generalization to data from different sources (of the same modality), and difficulties in ensuring the applicability of the predicted actions in the environment. Furthermore, the existing methods frequently are not developed to connect their predicted actions to specific locations in the environment, which is crucial for robot task and motion planning. In short, existing methods lack generalizability and environment-specific geometric grounding.


\begin{figure}
\centering
  \includegraphics[width=0.9\columnwidth]{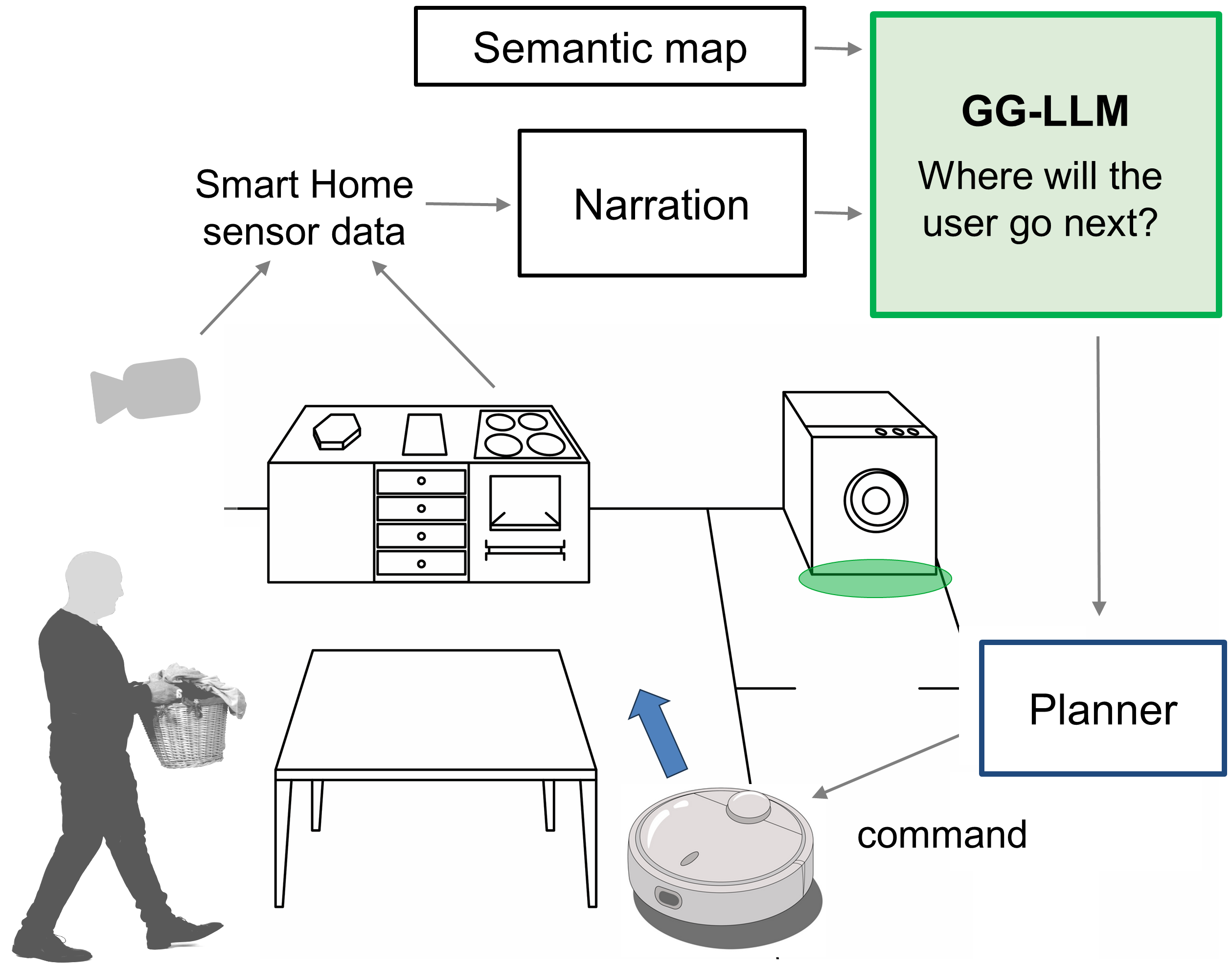}
  \vspace*{-5pt}
  \caption{We present a system that anticipates and localizes future human activities to enable human-aware robot task and motion planning. Our approach leverages the world knowledge contained within LLMs to infer probable future activities from observations of past actions. Using a semantic map, our system conditions the language model's symbolic predictions on the physical environment of the human, and localizes them within the environment for downstream robot planners.
  }
    \vspace*{-2em}
  \label{f:splash}
\end{figure}

We present an approach that addresses these challenges by leveraging large language models (LLMs) to infer a human's probable next actions while grounding these predictions in a semantic map of the environment (Fig.\ref{f:splash}). Our method utilizes natural language as a versatile interface to aggregate observations from a variety of input sources (including video or signals from connected appliances), ensuring adaptability to settings with diverse and variable sensing modalities. 
LLMs are trained on extensive text corpora depicting various human behaviors and therefore implicitly encode knowledge about human preferences and habits. We extract the foundational understanding of human activities from LLMs to anticipate the objects with which a human is likely to interact next without the need for additional training. We then ground these predictions in the physical environment of the human by connecting them to a semantic map of the scene, and use this information to build human-aware assistive robots.

We conduct simulated experiments in which a robot aims to achieve full room coverage while minimizing disruptions to humans in the environment. Our experiments encompass three distinct sequences of human activities, each executed across three diverse indoor environments. Our approach yields a significant reduction in disruptions to the human, thus demonstrating the benefit of incorporating localized activity predictions within a human-aware task planner.




In summary, our contributions are as follows:
\begin{itemize}
    \item We show that state-of-the-art LLMs encode information about common human behavioral patterns that can be used to predict human actions without further training.
    \item We present \ourmethod{} which can ground this information in a the physical world using a semantic map of the environment to effectively predict and localize future human activities.
    \item We demonstrate the utility of \ourmethod{} in human-aware robot planning in a floor plan coverage application. 
    We show that we can use \ourmethod{}'s predictions to reduce human disturbance by $29.2\%$ on average.
\end{itemize}
\commentout{
\begin{itemize}
    \item a new method that grounds LLM-based anticipation of future human action within the human's surroundings, thus localizing behaviors and ensuring admissibility of predicted actions
    \item integration of our method with human-aware task planning, showing a reduction of disturbances by 30%
    \item experimental validation (in simulation) of our method in which we show: robustness across environment and activities
    \item
    \item \vtxt{We present a novel evaluation methodology where we used a story-based human simulator...} \mtxt{I think our evaluation method is on the weaker side so I'd prefer to not emphasize it in the contribution statement}
\end{itemize}
}

\section{Related work}
\subsection{Large language and vision-language models in robotics}
Recent advances in LLMs have unlocked a variety of exciting capabilities in robotics. They have enabled the use of natural language as a flexible and intuitive communication layer, both between human operators and robots \cite{vemprala2023chatgpt,sharma2022correcting,ren2023robots,huang2023voxposer} and among cooperative multi-agent robot teams~\cite{zhang2023building}. They have further been applied in assistive robots that leverage the knowledge encoded in LLMs to rapidly understand and summarize human preferences from limited demonstrations~\cite{wu2023tidybot}. LLMs have also been proposed to independently execute robot planning tasks, but unresolved challenges remain in ensuring the certifiability and admissibility of the resulting plans~\cite{huang2022language}. The spatially-aware fusion of LLM embeddings with visual features has extended work on closed-vocabulary semantic maps (e.g.,~\cite{vasilopoulos2020reactive}) to yield robots capable of executing complex search and navigation tasks from open-vocabulary instructions~\cite{huang2023visual,gadre2022clip,kakodkar2023cartier}. Finally, vision-language models have been extended to predict robot actions based on visual and text inputs, allowing them to leverage training on internet-scale data for improved generalization in manipulation tasks~\cite{brohan2023rt}.

\section{Problem formulation}
\label{s:problem}
In this setting, a robot agent is executing a task (e.g., cleaning, inspection, or surveillance) that requires it to visit all $n_R$ rooms $R_i$ for $i \in {1,2,...,n_R}$ of an indoor environment $\mathcal{E}=\bigcup_{i=1}^{n_R} R_i$ while minimizing disturbances to humans present in $\mathcal{E}$. The total time horizon of this task planning problem is $T$. Without loss of generality, we assume that completing the task takes a constant time $T_{r}$ for each room and that once the robot enters a room, it stays there until task completion. 

We represent which room is occupied by the robot and human at time $t$ with one-hot vectors $\mathbf{x}_r(t)$ and $\mathbf{x}_h(t)$ (where $\mathbf{x}_r,\mathbf{x}_h \in \{0,1\}^{n_R}:\sum_{i=1}^{n_R} x_i(t)=1$). 

Disturbance to the human can mean many things: being within sight or earshot of the human, being closer than a certain distance to the human, or occupying the same room partition as the human. We define disturbance as same-room occupation. We introduce a binary variable $d_{rh}(t)$ that equals $1$ if the human is being disturbed by the robot at time $t$ and $0$ otherwise, and obtain $d_{rh}(t)=\mathbf{x}_r(t)\cdot \mathbf{x}_h(t)$ for our room-level definition of disturbance. The total number of disturbances $D_T$ over the given time horizon is $D_T = \sum_{t=0}^T d_{rh}(t)$ and the total time spent in each room by the robot is $\mathbf{X}_r=\sum_{t=0}^T \mathbf{x}_t(t)$.


The objective of the robot agent is to achieve full floor plan coverage (i.e., $ X \geq T_r~ \forall~ X \in \mathbf{X}_r$) while minimizing the total disturbances $D_T$. An additional desirable property of the robot task planner is to achieve full coverage quickly (i.e., minimizing the time $t_c$ required to achieve full coverage for the first time).

The human is executing a sequence of \textit{activities} of daily living (e.g., doing the dishes), where each activity consists of one or multiple \textit{atomic actions} (e.g., collecting dishes, loading the dishwasher, starting the dishwasher, ... ). This results in a sequence of human atomic actions $\mathbf{a}_h=[a_{h,0}, a_{h,1}, ..., a_{h,n}]$ with varying duration. 

If $\mathbf{a}_h$ is known and each atomic action implies a deterministic room occupation, we can obtain an optimal room sequence by solving the corresponding mixed integer linear program. In our scenario (as in most real-world applications), however, the series of human actions is not known, rendering the above planning approach infeasible. Instead, the robot's knowledge is limited to partial observations of the human's past atomic actions. Optionally, it may also receive information concerning the status of the apartment and the various network-enabled devices and appliances contained therein.

\section{Method}
\commentout{
\subsection{Approach}
\vtxt{I suggest we remove this first paragraph. Against any fixed policy (probabilistic or not), there is an optimal deterministic policy but computing it is not trivial.I can tell you more later.}
If the human were to follow a probabilistic policy in which they select a room to occupy uniformly at random from all available rooms at each time point, the optimal robot policy would be to either randomly select a room from all rooms at each transition point 
or to follow a fixed permutation of rooms.
}
In this section, we introduce \ourmethod: a system that improves upon policies that assume fully randomized human room occupation by leveraging the insight that past human actions exhibit correlations with future actions, and that human actions are intrinsically linked to the objects in their environment. In designing our system, we assume that the human's behavior is independent of the robot.

At the core of our system lies a symbolic-geometric reasoning module. This module first reasons about plausible sequences of actions at the symbolic level. It predicts probable next actions based on past actions and anticipates the potential items with which the human is likely to interact next. As effective planning for human-aware robot behavior necessitates environment-aware geometric reasoning, 
we next ground these symbolic predictions in the physical environment using a semantic map $\mathcal{M}$, which contains the bounding boxes and semantic labels for each room and object instance (multiple object instances can share the same semantic label). Our robotic agent uses these geometrically grounded predictions of future human actions to reduce disturbances to the human as it moves through the environment.

\subsection{System overview}

\begin{figure}
\centering
  \includegraphics[width=0.8\columnwidth]{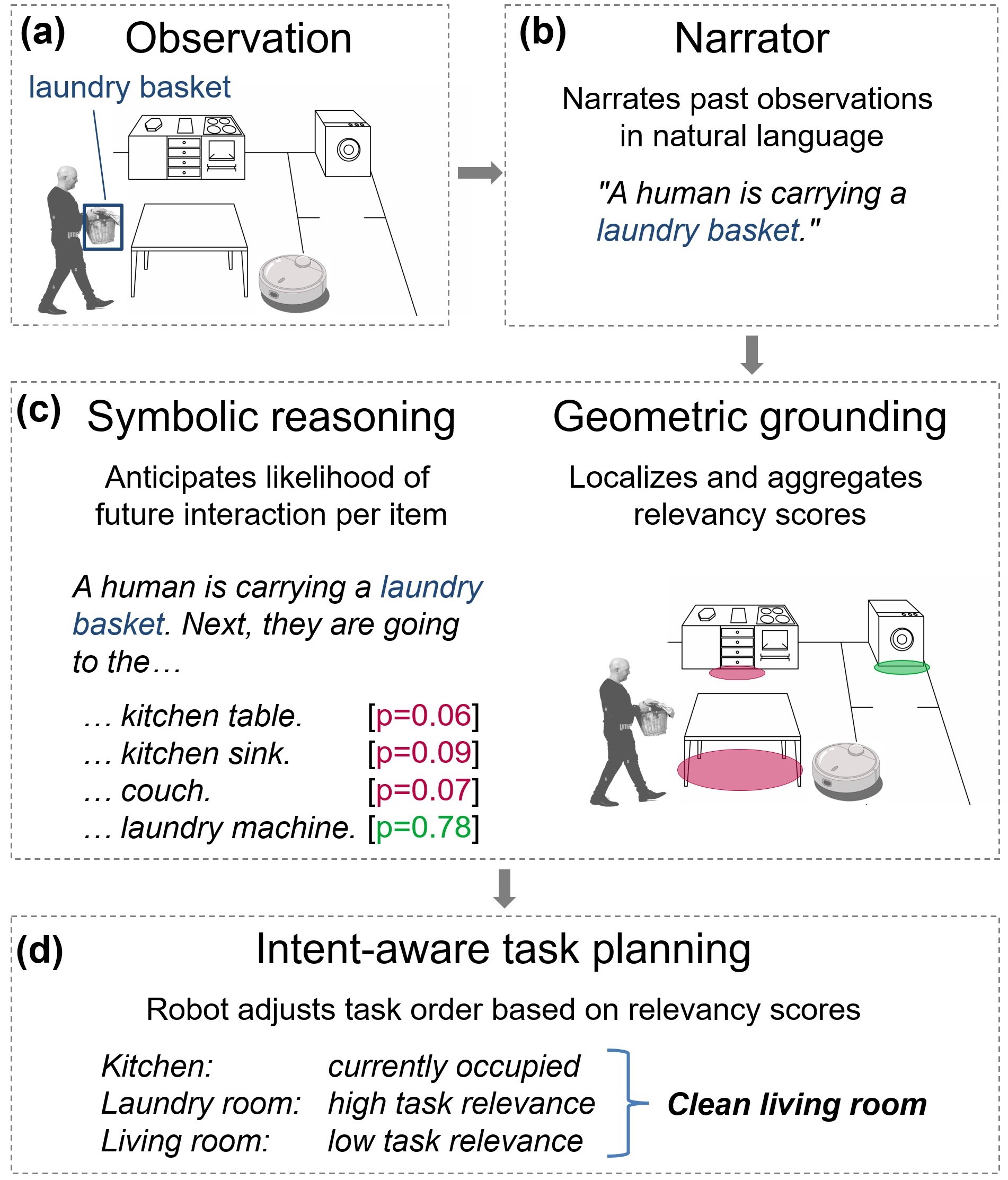}
  \vspace*{-5pt}
  \caption{System overview. Our system receives observations on past human activity from connected home appliances, cameras, and other sensors (a), and relies on a narrator module that translates these signals into a natural language narration of past human activities (b). This narration is passed into the symbolic-geometric reasoning module (c). This module leverages an LLM to predict future actions from the narration of past activities. We condition the LLM to predict items of future relevance with a forward-looking binding sequence (e.g., `Next, they are going to the ...') to the narration. We then use the LLM to score each item from the semantic map according to its likelihood of occurring in continuation of the prompt. We aggregate item-wise scores across map partitions to infer which partitions are most likely going to be occupied by the human. Finally, we demonstrate how a robot can use these geometrically-grounded predictions to minimize disturbances when planning its tasks (d).}
      \vspace*{-.5cm}
  \label{f:sys_arch}
\end{figure}

\ourmethod's system architecture, depicted in Fig.~\ref{f:sys_arch} consists of three sequential modules: a narrator module, a symbolic-geometric reasoning module, and an intent-aware planning module. The narrator module receives observations from various sources such as video feeds from robot or apartment cameras, and data streams from connected devices and appliances, aggregates these observations, and creates a natural language description of the human's past actions and the apartment state. The symbolic-geometric reasoning module comprises two submodules. The symbolic reasoning submodule analyzes past actions and assigns a score to each object within the environment, which reflects the likelihood of human interaction with said object in the near future. The geometric grounding submodule localizes these object scores within the semantic map and aggregates scores across different subsections of the scene, where a subsection can consist of a room partition, a full room, or multiple rooms. This process yields an aggregate score for each subsection which is proportional to the predicted probability of future human occupancy of said subsection. Finally, the intent-aware planning module leverages the spatially aggregated scores generated by the symbolic-geometric reasoning module to decide in which subsection the robot should operate next. In the following subsections, we provide a detailed description of each of the modules. We also provide a formal definition of the semantic map $\mathcal{M}$, as it is a crucial component of the symbolic-geometric reasoning module. 


\subsection{Narrator}
At each time point, the narrator module receives a set of observations $\mathbf{o}(t)$ from the robot sensors and network-enabled appliances. It aggregates these observations into an observation history object $\mathbf{h}(t)=\{\mathbf{o}(t-n_h),...,\mathbf{o}(t)\}$ that includes the observations of the current and last $n_h$ time steps. It then transforms the history object into a natural language narration of the history $\mathbf{N}_h$. This narrator module can leverage generic image (e.g.,~\cite{liu2023visual}) or video captioning models (e.g.,~\cite{zhao2023learning, video_narration_2022}), or models trained specifically to detect human activities (e.g.,~\cite{rajasegaran2023benefits}) to transcribe camera data to natural language, or rely on activity detection based on mobile devices or smart homes. 
In our experiments, we assume that a description of previous human actions is available. 

\subsection{Semantic map}
The semantic map $\mathcal{M}$ of our scene is a (closed-vocabulary) collection of $i$ items $i_i \in I$, where each item $i_i$ is associated with a semantic label $l_j \in L$, its position $p_i$, and its bounding box $b_i$. Multiple items can have the same semantic label (e.g., there can be multiple item instances of type \textit{mug} in a scene), and bounding boxes can be overlapping. The map also includes room-level bounding boxes (Fig.~\ref{f:geo_grounding_draft}(a)).

\subsection{Symbolic reasoning}

The symbolic reasoning submodule projects with which objects the user is likely interacting next. It takes as input the language narration of the history object $\mathbf{N}_h$, as well as a list of items available in the environment (which can be extracted from the semantic map $\mathcal{M}$). It assigns a relevancy score $s_i$ to each of the items in $\mathcal{M}$. 

While one could learn the mapping from past actions to relevancy scores for each item, this approach would require extensive data and computational power due to the large variability in human activities and preferences.
We overcome this challenge by leveraging the world knowledge encoded within LLMs, which allows us to achieve zero-shot symbolic reasoning about human behaviors.

We take the following approach to extract relevant world knowledge from the language model: First, we construct a prompt that combines the history narration  $\mathbf{N}_h$ with a binding sequence $\mathbf{B}$ designed to induce the language model to reason about what the human will do next (we use $\mathbf{B}=\text{`Next, the human will go to the '}$ in our experiments). We then generate a set of possible completions $C=\{\mathbf{c}_i\}$ of the prompt $\mathbf{N}_h+\mathbf{B}$ (where $+$ indicates string concatenation). While one could rely on language models to generate $C$, it can be challenging to ensure in a principled way that this approach yields a $C$ which covers all relevant items and room sections within the environment. Therefore, we generate $C$ by extracting the semantic labels $l_j$ of all items from $\mathcal{M}$. An illustrative example is shown in Fig.~\ref{f:sys_arch}.

We compute the relevancy score for each completion $c_i$ based on the total probability that a language model $M$ assigns to it when conditioned on  narration $\mathbf{N_h}$ and binding sequence $\mathbf{B}$. For this step, we first obtain the (model-specific) tokenizations of $\mathbf{N}_h$, $\mathbf{B}$, and $c_i$, resulting in token sequences $\bar{\mathbf{N}}_h$, $\bar{\mathbf{B}}$, and $\bar{c}_i^{1:l_i}$. It is worth noting that $\bar{c}_i^{1:l_i}$ is a sequence of length $l_i$, where $l_i$ varies across different $c_i$. Now, let $p_M(z_{n+1} | z_1, . . . , z_n)$ be the probability that the language model $M$ assigns to token $z_{n+1}$ as a continuation of the sequence $z_1, . . . , z_n$. We compute the relevancy score for completion $c_i$, $s_{c,i}$ as shown in Eq.~\ref{e:rel_score}.
\begin{equation}
\label{e:rel_score}
s_{c,i} = \prod_{j=1}^{l_i}
p_M(\bar{c}_i^j|\bar{\mathbf{N}}+\bar{\mathbf{B}}+\bar{c}_i^{1:j-1})
\end{equation}

Finally, we assign an item-level relevancy score $s_i$ to each item $i_i$ in the semantic map $\mathcal{M}$ as $s_i=s_{c,i}/n_{l,i}$, where $n_{l,i}$ is the number of occurrences of the semantic label associated with item $i_i$. 

\subsection{Geometric grounding}
\begin{figure}
  \includegraphics[width=\columnwidth]{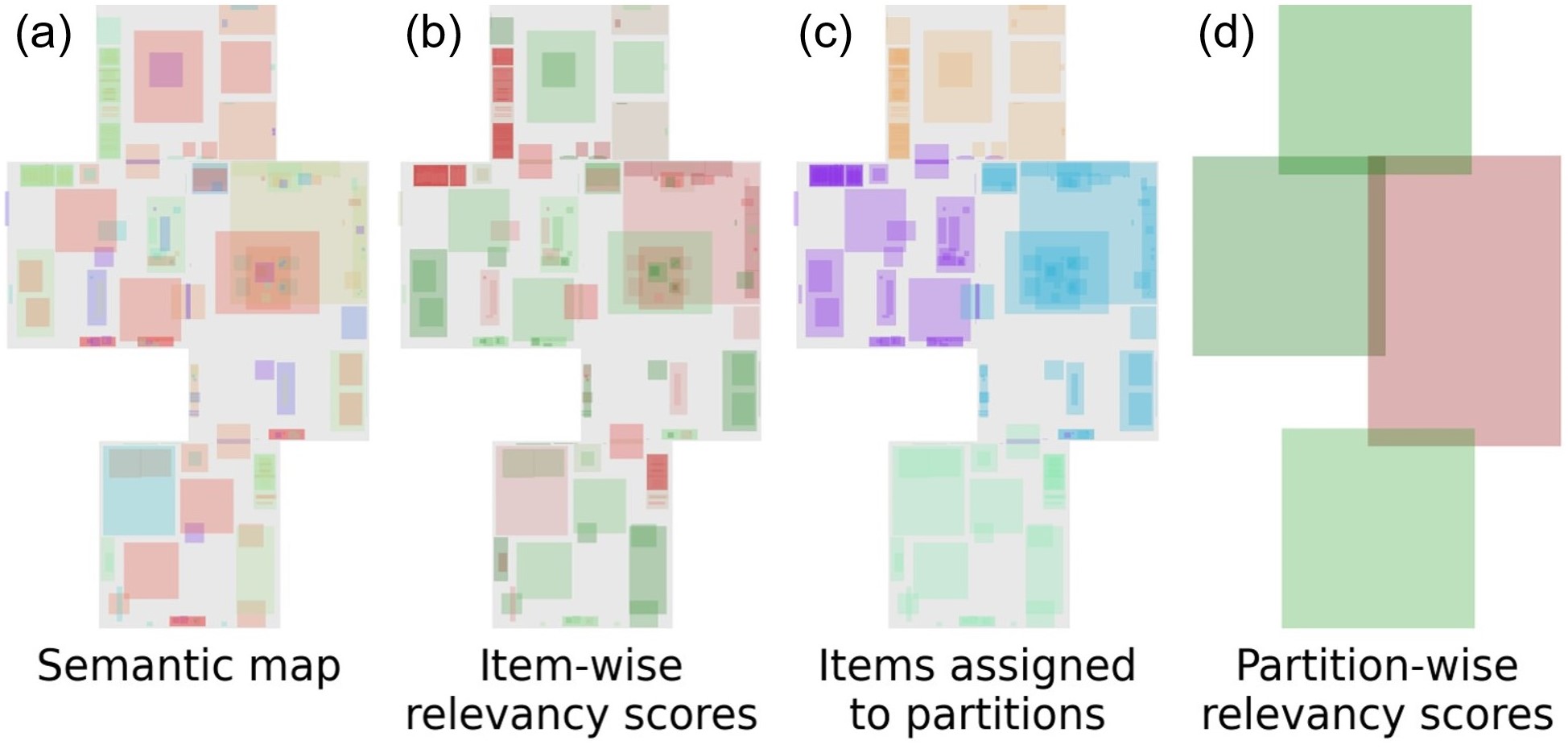}
  \vspace*{-15pt}
  \caption{The geometric grounding submodule relies on a semantic map $\mathcal{M}$ (a) to localize the predictions from the symbolic reasoning module (item color corresponds to item type). It does so by assigning a relevancy score (computed by the symbolic reasoning submodule) to each item in the semantic map as shown in (b) (item color corresponds to relevancy score). Each of the items is then assigned to the spatial map partition with which its bounding box has the larges overlap. We show an exemplary assignment where the map is partitioned at the room level in (c), where the item color reflects partition assignments. Finally, the item-wise relevancy scores are summed across each partition to obtain partition-wise relevancy scores for downstream robot planning tasks (d).
  }
      \vspace*{-1.5em}
  \label{f:geo_grounding_draft}
\end{figure}

The geometric grounding submodule receives the relevancy scores $s_i$ for each item in the semantic map $\mathcal{M}$ and aggregates these scores across different spatial subdivisions of the scene as illustrated in Fig.~\ref{f:geo_grounding_draft}. These subdivisions may encompass room partitions, entire rooms, or even multiple rooms. The output of the geometric grounding submodule is a partition-level relevancy score $\bar{S}_j$ for the $j$-th spatial partition $P_j$. We obtain $\bar{S}_j$ by first assigning each item to the subdivision with which its bounding box $b_i$ has the largest overlap, and then calculate the partition-level relevancy score as the sum of the item-wise relevancy scores of all items within a given partition. Let $V_o(b_i,P_j)$ be the overlap volume between bounding box $b_i$ and partition $P_j$, and $I(\cdot)$ the indicator function. Then, we can compute the the partition-level relevancy score as follows: 
\begin{equation}
    \bar{S}_j = \sum_{i} s_i\cdot I(\arg \max_k V_o(b_i,P_k)=j)
\end{equation}

\subsection{Intent-aware planning}
\label{s:planning_strategies}
The intent-aware planning module defines a policy $\pi(\bar{S})$ which selects the future partition $FP_j$ to which the robot should move to as a function of the set of partition-level relevancy scores $\bar{S}=\{\bar{S}_j\}$. In our experiments in Section\ref{s:exp}, we compare three different policies:
\subsubsection{Naive}
The naive policy $\pi_N(\bar{S})$ selects the next partition to tend to uniformly at random:

\begin{equation}
FP\sim\text{Uniform}[\{P_k\}]
\end{equation}

\subsubsection{Greedy avoidance}
The greedy avoidance policy $\pi_{GA}(\bar{S})$ always chooses the partition with the lowest relevancy score (ties are broken at random):
%
\begin{equation}
FP=P_k: k = \arg\min_i \bar{S}_i
\end{equation}

\subsubsection{Informed avoidance}
The informed avoidance policy $\pi_{IA}(\bar{S})$ discards the partition with the highest relevancy score and  selects from the remaining partitions uniformly at random:
%
%
\begin{equation}
FP\sim\text{Uniform}[\{P_k: k \neq \arg\max_i \bar{S}_i \}]
\end{equation}
 

\section{Human activity simulation}
\label{s:vhome}
We use the VirtualHome simulator~\cite{puig2018virtualhome}, which was developed by Puig et al. (2018) for the simulation of household activities. In this section, we review relevant features of the simulator and introduce the environments used in our experiments. The VirtualHome simulator leverages the Unity3D game engine to animate one or multiple characters. Given an indoor scene and an activity program (i.e., a series of atomic actions), the simulator returns the character's pose over time, and optionally renderings from cameras inside the environment.

\begin{figure}
  \includegraphics[width=\columnwidth]{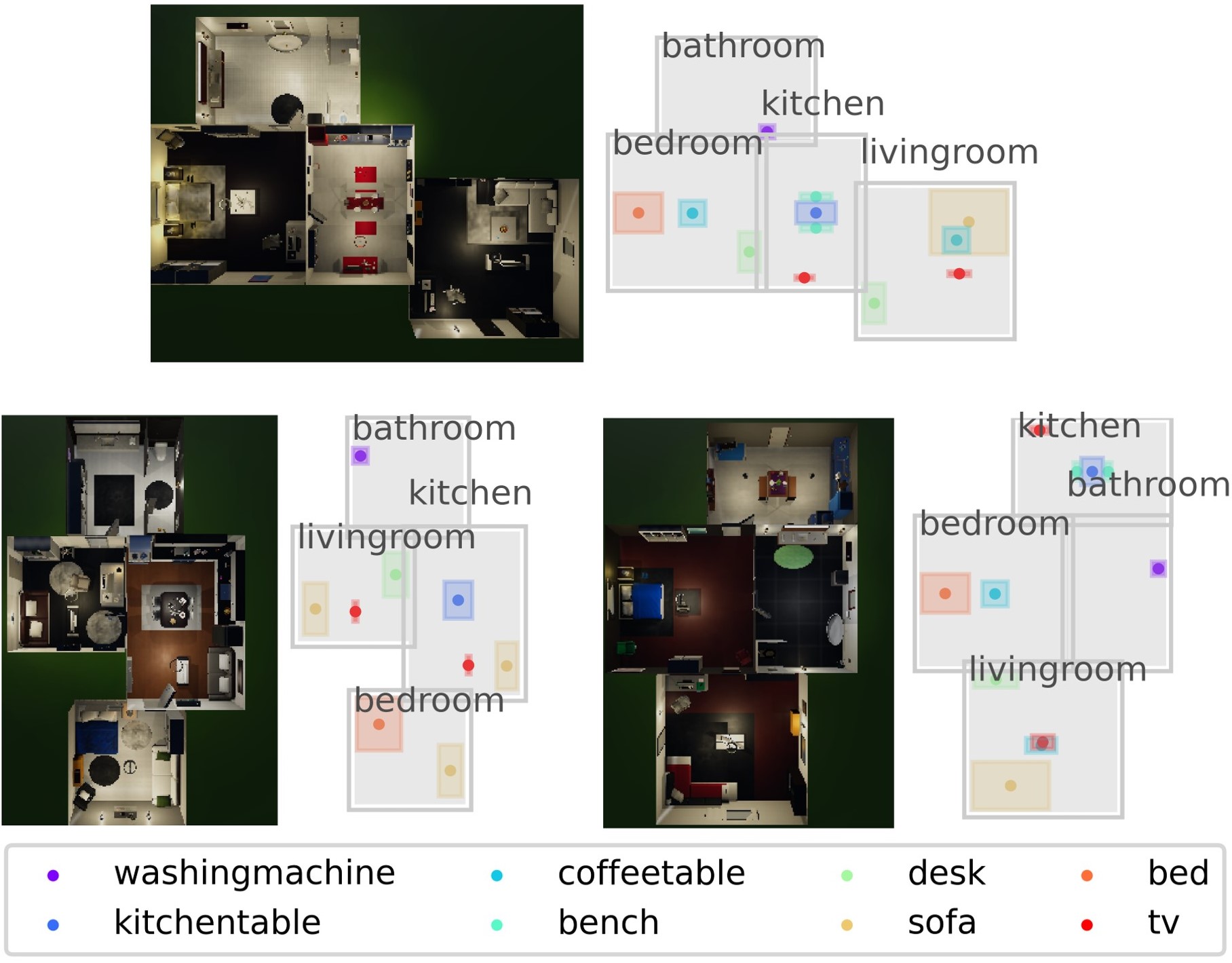}
  \vspace*{-18pt}
  \caption{Top-down view of of our environments (left) and the corresponding semantic maps. Only a small subset of the items present is shown in the map visualization for readability.}
      \vspace*{-1.5em}
  \label{f:scene_maps}
\end{figure}

The VirtualHome simulator provides a range of indoor environments with varying room layouts and item availability. We select three of these environments for our validation experiments; their rendered top down view and a semantic map including key items are shown in Fig.~\ref{f:scene_maps}. All three environments consist of three rooms. Environment $\mathcal{E}_0$ contains 115 item classes and 443 total items. Environments $\mathcal{E}_1$ (and $\mathcal{E}_2$) contain 98 (100) item classes, and 357 (324) total items.

\begin{table*}[t]
\centering
\caption{Prompt variation experiment results}
      \vspace*{-0.15cm}
\begin{tabular}{cccccc}
\textbf{Planning policy} & \textbf{Prompt}        & \textbf{Narration} \bm{$\mathbf{N}_h$} & \textbf{Binding sequence }\bm{$\mathbf{B}$} & \bm{$D_T$}  & \bm{$F$~{[}\%{]}}\\ 
 \midrule[1.25pt]
Random: $\pi_{N}$ & - & -  & -  &       317.1±86.3       & 0.2  \vspace{2pt}
      \\
\multirow{6}{*}{GG-LLM: $\pi_{IA}$} & $\mathbf{P_0}$ & activity history  & ``Next, they will go to the "  &  163.4±59.0 &           0.7         \\
&$\mathbf{P_1}$ & activity history  & ``The next object they are walking to is the "   & 163.4±59.0  &  0.7  \\
&$\mathbf{P_2}$ & activity history  & ``After this, they are going to interact with the " &  237.2±72.9  & 2.0 \\
&$\mathbf{P_3}$ & activity history  & ``The most tasty object in the world is the "      &  316.0±85.7 & 25.1 \\
&$\mathbf{P_4}$ & none              & ``The most expensive object in the world is the "   & 285.3±79.1  & 75.3  \\
&$\mathbf{P_5}$ & none              & ``Next, they will go to the "           &     285.3±79.1   &   75.3     \\
\end{tabular}
\label{t:prompt_variation}
      \vspace*{-1.5em}
\end{table*}

Characters in the VirtualHome simulator can be instructed to execute an \textit{activity program}, which consists of a list of atomic actions. Our validation relies on three different activity programs describing common household activities; they consist of 18, 16, and 42 atomic actions, respectively.
Atomic actions in the VirtualHome simulator are encoded in the following simulator-specific format:
$$
[\text{action}]~\langle l_j \rangle (ID_j)~ ...
$$
Here, $l_j$ refers the semantic label of the item (e.g., 'mug') and $ID_j$ to the identifier of the item (as there can be multiple instances of each item within a program). An action can be associated with zero, one, or multiple items.

The human will consistently refer to the same item in simulation for items which are listed to have the same identifier in a program. For example, when simulating the activity program below, the character will select an arbitrary mug from all available mugs, walk towards it, and then grab this particular mug. 
\begin{equation*}
  \begin{gathered}
[\text{Walk}]~\langle \text{mug} \rangle~ (\text{1})
\\
[\text{Grab}]~\langle \text{mug} \rangle ~(\text{1})
\end{gathered}
\end{equation*}

If the second line read `$[\text{Grab}]~\langle \text{mug} \rangle~ (\text{2})$' instead, the character would aim to pick up a different mug than the one it walked to. The item identifier does not provide a mapping to a pre-determined item in the simulation, leaving the decision of which items to interact with to the simulated human agent. The simulator also handles path planning and inverse kinematics for character animation. 



\begin{figure*}[]
  \includegraphics[width=\textwidth]{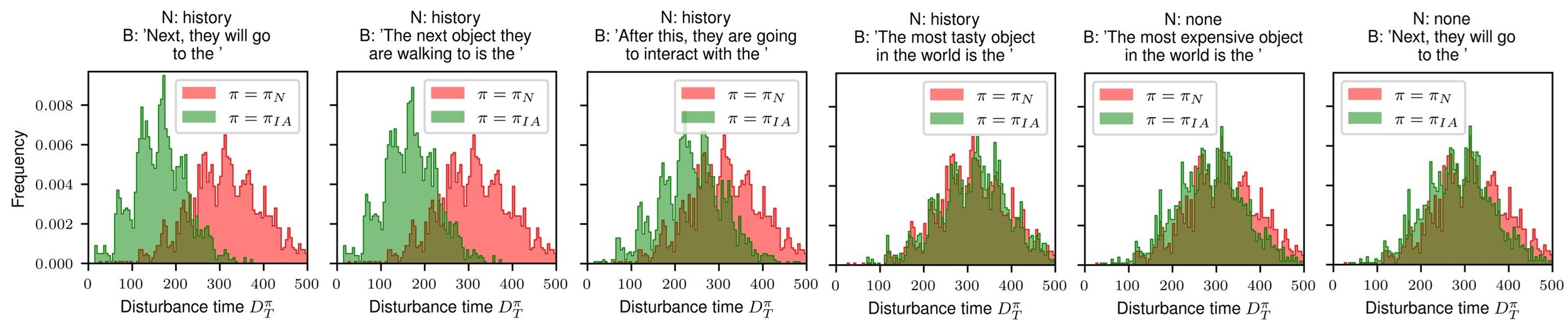}    \vspace*{-20pt}
  \caption{Histograms of disturbance times for random and intent-aware policies across different narrations $\mathbf{N_h}$ and binding sequences $\mathbf{B}$. When the prompt contains both a narration of past human actions and a binding sequence cuing the anticipation of future actions, our approach successfully reduces the number of disturbances (a-c). If we employ a narration which does not relate to past human activity or a binding sequence that has no semantic relation to future human activities, the behavior of the LLM-informed agent closely resembles the random agent.}
      \vspace*{-1.em}
  \label{f:results_raw}
\end{figure*}

\section{Experiments}
\label{s:exp}
We validate whether our zero-shot intent-aware planning approach is able to reduce the number of human disturbances during coverage through extensive experiments. We rely on the third-party simulator described in Section~\ref{s:vhome} to simulate human activities for an unbiased evaluation of our method. 
We investigate the effect of prompt construction on our method in Section~\ref{s:prompt_var}, and study its robustness to environment and activity variations in Section~\ref{s:env_var}.

In all our experiments, we extract the information needed to build the semantic map from the VirtualHome environment description and employ the LLaMA2~\cite{touvron2023llama} as the language model in our symbolic reasoning submodule. The narrator module has access to the list of all previously executed atomic actions; each atomic action is only observable upon its full completion in simulation.

To accurately capture the behavior of the probabilistic policies, we execute 1000 restarts of the robot policy simulation for each setting and policy (where the robot's position is initialized in one of the rooms uniformly at random). For each restart, we record the total disturbance $D_T$; the time required to achieve full coverage $t_c$; and the number of runs in which the robot fails to achieve full coverage ($F$).

\subsection{Prompt variation}
\label{s:prompt_var}
In this experiment, we investigate the effect of different prompt formulations on our system's ability to reduce the number of disturbances to the human. We compare the six prompts shown in Table~\ref{t:prompt_variation}, which exhibit variations in the narration $\mathbf{N_h}$ and binding sequence $\mathbf{B}$. To study the effect of $\mathbf{N_h}$, we compare narrations that describe the activity history in prompts $\mathbf{P_{0}}$, ..., $\mathbf{P_{3}}$ with blank narrations which do not condition the answer on any observations in prompts $\mathbf{P_{4}}$, $\mathbf{P_{5}}$. We further evaluate three binding sequences designed to cue the language model to reason about future human activities ($\mathbf{P_{0}}$, $\mathbf{P_{1}}$, $\mathbf{P_{2}}$) and two binding sequences that are unrelated to human activity ($\mathbf{P_{3}}$, $\mathbf{P_{4}}$). We conduct this prompt variation study on one environment for one human activity program. We run $1000$ restarts of the robot task with room times $T_r=25$ and $T_r=50$ (2000 runs total).

We report the mean (± standard deviation) of the disturbance time $D_T$ and the number of runs that failed to achieve full coverage ($F$) across all restarts for each prompt in Table~\ref{t:prompt_variation}. To provide a complete picture of the disturbance distribution over these runs, we also report the histograms of the resulting disturbances $D_T$ for each prompt in Fig.~\ref{f:results_raw}.

We compare the LLM-informed policies to the baseline random policy $\pi_N$. We observe that combining a narration of past activities with a binding sequence that cues the prediction of future human actions ($\mathbf{P}_0$, $\mathbf{P}_1$, $\mathbf{P}_2$) significantly reduces the number of collisions. Our method shows robustness to the variations in the binding sequence we investigated as long as the binding sequence guides the language model towards future human behaviors ($\mathbf{P}_0$, $\mathbf{P}_1$, $\mathbf{P}_2$).

We further observe that our LLM-based policy performs similar to the naive policy when it is either cued with a binding sequence that is not related to human activity  ($\mathbf{P}_3$, $\mathbf{P}_4$) or provided no information on past human actions  ($\mathbf{P}_4$, $\mathbf{P}_5$). This confirms that our approach successfully extracts inherent world knowledge about human behavior patterns while also tying its predictions to past observations and admissible environment interactions. 





\subsection{Environment and activity variation}
\label{s:env_var}

We conduct experiments on three different environments (visualized in Fig~\ref{f:scene_maps}) and run three activity programs in each of them. We study the efficacy of the policies we introduce for our intent-aware planning module in Section~\ref{s:planning_strategies} -- naive ($\pi_N$), greedy avoidance ($\pi_{GA}$), and informed avoidance ($\pi_{IA}$) -- for three distinct values of the time $T_r$ required to complete the task in a given room (25, 50, and 100).

\begin{figure}
  \includegraphics[width=\columnwidth]{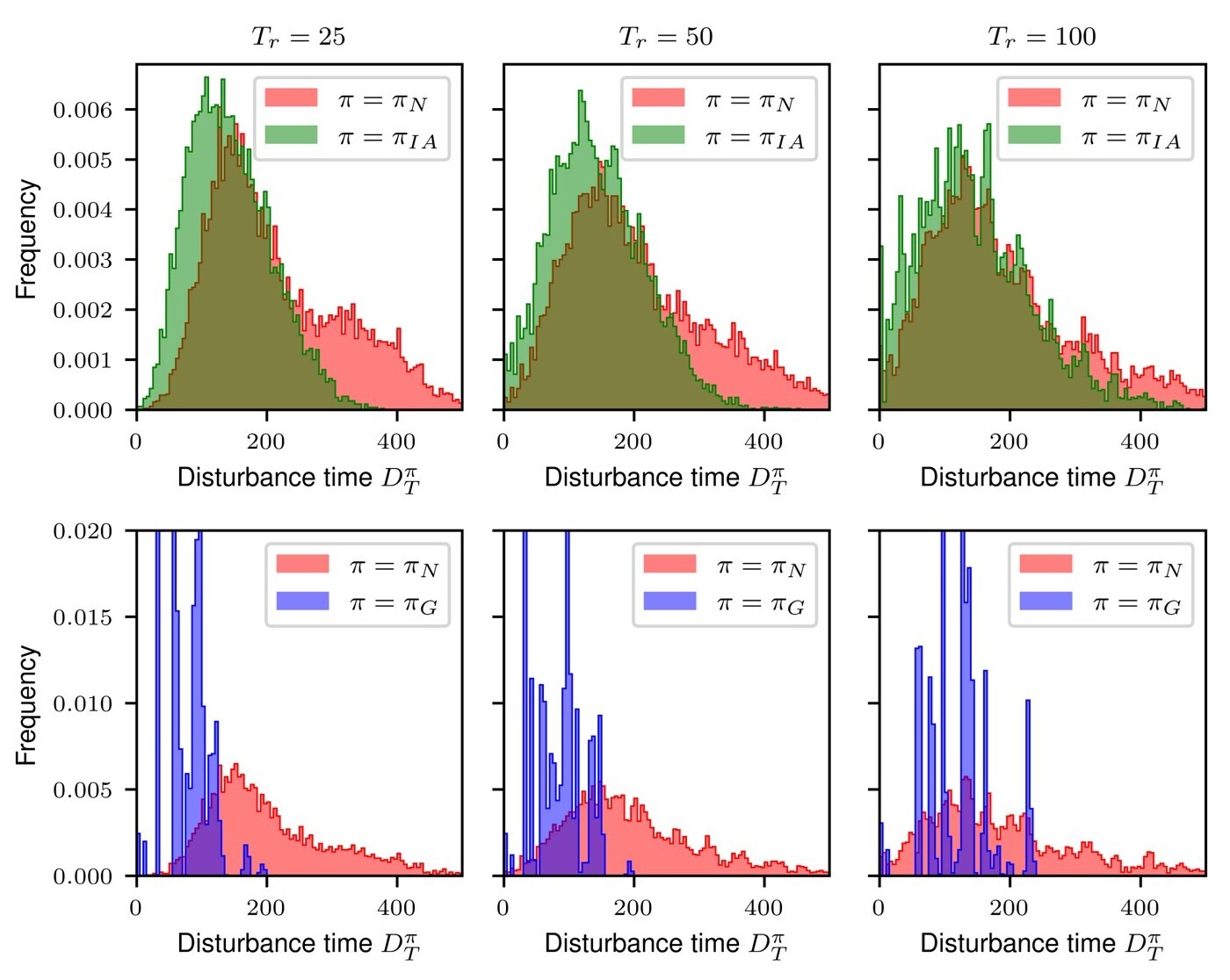}
    \vspace*{-20pt}
\caption{
  Histograms of disturbance times for random and intent-aware policies across three different environments, three activity programs, and three values for the time $T_r$ the robot stays in each room. The GG-LLM policies ($\pi_{IA}$, $\pi_{GA}$) achieve lower disturbances on average across all scenarios than the random policy $\pi_N$. The difference is least pronounced for the largest value of $T_r$. For readability and y-axis scale compatibility, we show $\pi_{IA}$ and $\pi_{GA}$ on different panels.}
      \vspace*{-1em}
  \label{f:results_raw_full_exp}
\end{figure}

The histograms of the resulting disturbances $D_T$ obtained across all environments and all activity programs are shown in Fig.~\ref{f:results_raw_full_exp}, shown separately for each room time $T_r$. The mean disturbance time, time to coverage, and percentage of failed runs (averaged across all environments,  activity programs, and room times $T_r$) achieved with each of the policies are shown in Table~\ref{t:overall_performance}. On average, the disturbance incurred by the intent-aware policy $\pi_{IA}$ is $29.2\%$ lower when comparing to the baseline $\pi_{N}$. This gain comes at the cost of only a minor increase in runs for which coverage was not achieved.

\begin{table}[h!]
\caption{Overall performance comparison}
      \vspace*{-0.15cm}
\begin{tabular}{lccc}
\textbf{Planning policy}     & \bm{$D_T$} & \bm{$t_c$} & \bm{$F$ ~{[}\%{]}} \\ 
\midrule[1.5pt]
Random: $\pi_{N}$  &  205.1±114.3      &   355.3±225.2     &     18.3         \\
GG-LLM: $\pi_{IA}$ &  145.3±74.3     &  428.8±264.9     &     20.5         \\
GG-LLM: $\pi_{GA}$ &    100.6±50.0   &  251.6±124.8  &   93.3  \\
\end{tabular}
      \vspace*{-0.25cm}
\label{t:overall_performance}
\end{table}



The greedy avoidance policy $\pi_{GA}$ incurred a smaller, equal, and larger number of disturbances than the naive policy $\pi_{N}$ in $84.8\%$, $0.7\%$, and $14.5\%$ of the restarts. Its time to achieve full coverage was smaller, equal, and larger in $0.7\%$, $17.8\%$, and $81.5\%$ of the restarts. The informed-avoidance policy $\pi_{IA}$ incurred a smaller, equal, and larger number of disturbances than the naive policy $\pi_{N}$ in $70.3\%$, $0.6\%$, and $29.0\%$ of the restarts, respectively. Its time to achieve full coverage was smaller, equal, and larger in $33.0\%$, $17.5\%$, and $49.5\%$ of the restarts. The naive, greedy-avoidance, and informed-avoidance policy failed to achieve full coverage in $17.9\%$, $93.3\%$, and $20.8\%$ of restarts. 

We observe that both GG-LLM policies (i.e., informed avoidance $\pi_{IA}$ and greedy-avoidance $\pi_{GA}$) lead to a significant reduction in disturbance across all environments, activity programs, and room times $T_r$.  
The benefit of our approach is least pronounced for the largest room time we explored ($T_r$). This is expected: given infinite restarts, randomized starting rooms, and $T_r$ approach total time horizon $T$, the average performance across restarts for policies $\pi_N$ and $\pi_{IA}$ converges to the same value.

While the greedy-avoidance policy $\pi_{GA}$ - always selecting the room least likely to be occupied by the human - results in even fewer disturbances than the informed-avoidance policy $\pi_{IA}$, it fails to achieve full coverage far more frequently. 
The results highlight the utility of our geometrically grounded prediction method in human-aware task planning.



\section{Conclusion}
We presented \ourmethod{}, a method that leverages the world knowledge embedded within state-of-the-art LLMs to anticipate upcoming human actions based on narrations of past activities. In \ourmethod{}, a geometrical grounding step ties LLM predictions to objects in the environment through a semantic map. Our method thus provides robots with a semantic and spatial understanding of future human actions. We demonstrate how they can use this understanding in a human-aware coverage planning task, reducing disturbances to the human by $29.2\%$. 

While we demonstrate the potential of \ourmethod{}, there are a number of exciting directions to expand upon this work. These include additional ablation studies with different language models and prompt formulations, the extension of \ourmethod{} to work with open-vocabulary visual-semantic maps, and the incorporation of temporal reasoning to enable high-resolution predictions of human trajectories for smooth cooperation. Overall, we anticipate that our approach will not only enhance the capabilities of human-aware service robots but also open up new possibilities for active and effective human-robot collaboration in a wide range of settings, such as homes, hospitals, and beyond.





\balance
\bibliographystyle{IEEEtran}
\bibliography{IEEEabrv,references}

\end{document}